\pdfoutput=1

\documentclass[11pt]{article}

\usepackage[]{acl2025}

\usepackage{times}
\usepackage{latexsym}

\usepackage[T1]{fontenc}

\usepackage{microtype}

\usepackage{inconsolata}

\usepackage[utf8]{inputenc}
\usepackage{booktabs,multirow}
\usepackage{enumitem}
\usepackage{kotex} 
\usepackage{color}
\usepackage{rotating}
\usepackage{amsmath}
\usepackage{amsfonts}
\usepackage{pifont}
\usepackage{array}
\usepackage{amssymb}
\usepackage{algpseudocode}
\usepackage{algorithm}
\usepackage{float}
\usepackage{tcolorbox}
\usepackage{multicol}
\usepackage{lipsum}
\usepackage[normalem]{ulem}
\usepackage{soul}
\usepackage{listings}
\usepackage{amsmath} 
\usepackage{adjustbox}
\usepackage{tabularx}
\usepackage{booktabs}
\usepackage{multirow}
\usepackage{tabularx}
\usepackage{colortbl}
\definecolor{customhighlight}{HTML}{DDB6E4}
\sethlcolor{customhighlight}

\colorlet{instructionbg}{gray!15}
\colorlet{questionbg}{gray!25}

\newcommand{\cmark}{\textcolor{green!70!black}{\ding{51}}} 
\newcommand{\xmark}{\textcolor{red}{\ding{55}}}

\newcommand{\ie}{{\it i.e.}}
\newcommand{\eg}{{\it e.g.}}

\newcommand{\ours}{CARE}

\newcommand{\es}{\textcolor{orange}}

\title{Conflict-Aware Soft Prompting for Retrieval-Augmented Generation}

\author{Eunseong Choi, June Park, Hyeri Lee, Jongwuk Lee\thanks{\ \ Corresponding author} \\
        Sungkyunkwan University, Republic of Korea\\  
        \texttt{\{eunseong, pj00515, bluepig94, jongwuklee\}@skku.edu}}

\begin{document}
\maketitle
\begin{abstract}

Retrieval-augmented generation (RAG) enhances the capabilities of large language models (LLMs) by incorporating external knowledge into their input prompts. However, when the retrieved context contradicts the LLM’s parametric knowledge, it often fails to resolve the conflict between incorrect external context and correct parametric knowledge, known as \emph{context-memory conflict}. To tackle this problem, we introduce \emph{\textbf{C}onflict-\textbf{A}ware \textbf{RE}trieval-Augmented Generation (\textbf{\ours})}, consisting of a \emph{context assessor} and a \emph{base LLM}. The context assessor encodes external context into compact \textit{memory embeddings}. Through \emph{grounded/adversarial soft prompting}, the context assessor is trained to discern unreliable context and capture a guidance signal that directs reasoning toward the more reliable knowledge source. Extensive experiments show that \ours\ effectively mitigates context-memory conflicts, leading to an average performance gain of 5.0\% on QA and fact-checking benchmarks, establishing a promising direction for trustworthy and adaptive RAG systems\footnote{https://github.com/eunseongc/CARE}.

\end{abstract}

\section{Introduction}\label{sec:introduction}

\begin{figure}[t]
\includegraphics[width=1.0\linewidth]{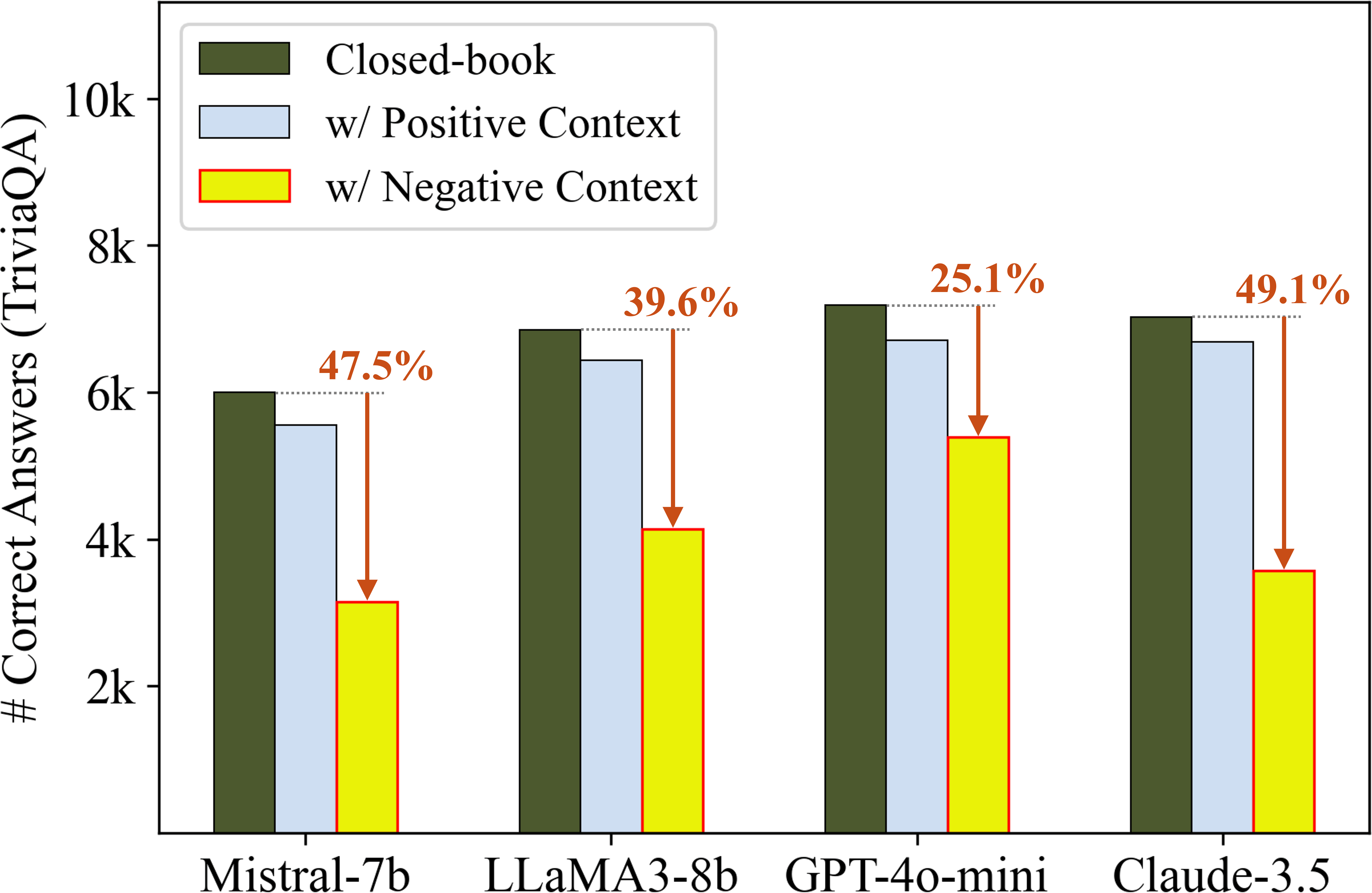}
\caption{LLMs struggle to resolve context-memory conflict. Green bars show the number of questions correctly answered without retrieval in a closed-book setting. Blue and yellow bars show performance when provided with a positive or negative context, respectively.}\label{fig:fig_llms_drop}
\vskip -0.1in
\end{figure}



Retrieval-augmented generation (RAG) serves as an effective strategy for enhancing large language models (LLMs) by grounding them in external information~\cite{nips/LewisPPPKGKLYR020/RAG, corr/RAG-survey1, corr/RAG-suvey2}. However, when the retrieved context contradicts the LLM's internal knowledge, it causes a critical vulnerability: the challenge of resolving the conflict between incorrect external context and correct internal knowledge, known as \emph{context-memory conflict}.

This challenge is exacerbated when incorrect yet highly ranked contexts serve as hard negatives. Conventional RAG, \ie, simply appending retrieved context to the prompt, struggles to discriminate between incorrect external context and correct parametric knowledge~\cite{coling/RenWQZ00W025/priori}. This misalignment leads to overriding correct internal representations, resulting in substantial performance degradation on questions that the model initially answered correctly. As shown in Figure~\ref{fig:fig_llms_drop}, we observed significant performance drops of 25.1-49.1\% across state-of-the-art LLMs when negative contexts were added to questions for which the LLM had already generated the correct answer without external context.

\begin{figure}[t]
\includegraphics[width=1.0\linewidth]{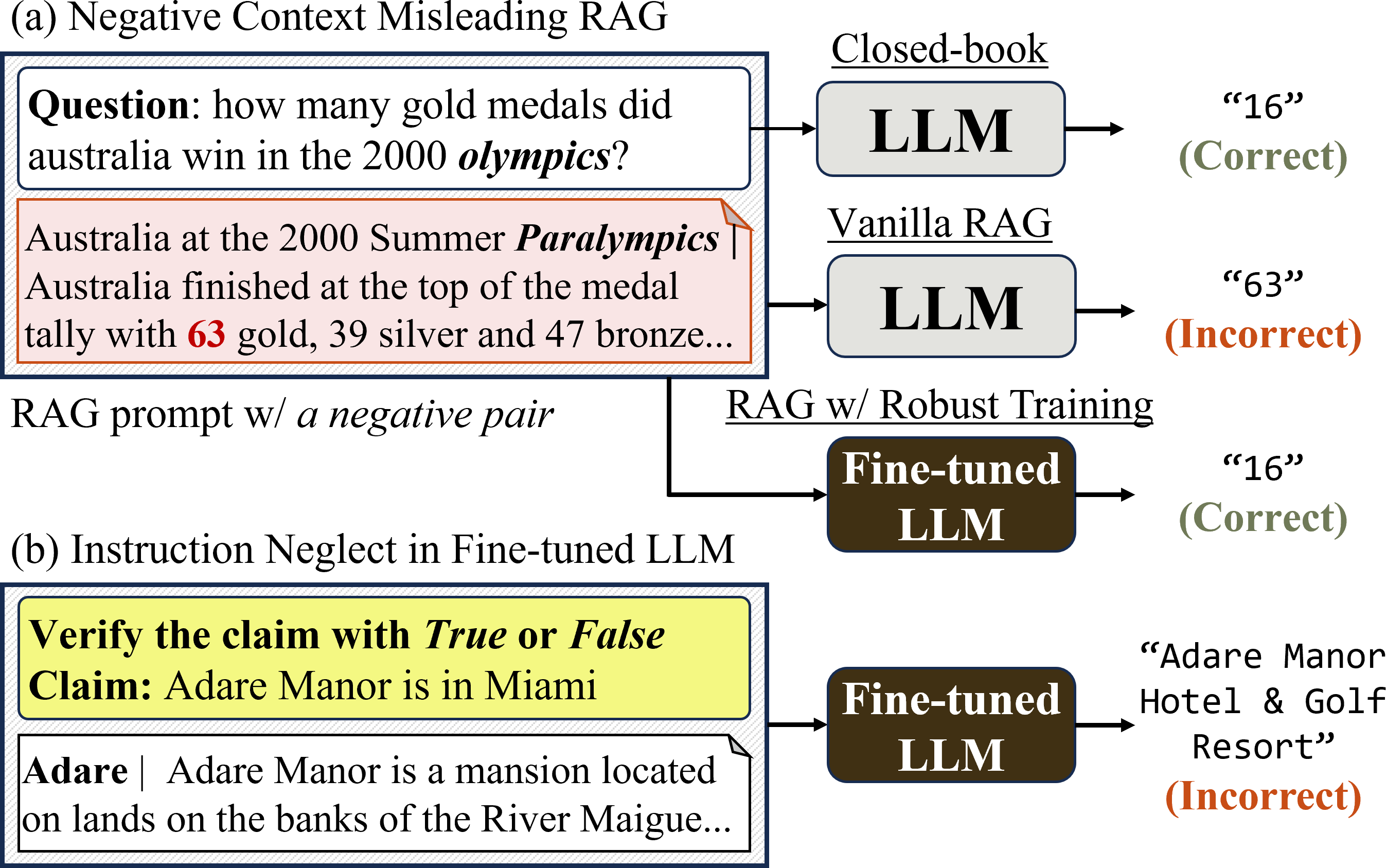}
\caption{(a) Vanilla RAG fails with negative context despite having the correct information in parametric knowledge. (b) Robust training can cause the LLM to disregard instructions and suffer from catastrophic forgetting, particularly when transitioning between tasks like question answering and fact verification.}\label{fig:fig_motivation}
\vskip -0.2in
\end{figure}

To mitigate context-memory conflict, existing studies such as adaptive retrieval~\cite{coling/RenWQZ00W025/priori, emnlp/BaekCKLL25/ProbingRAG} and the decoding strategies~\cite{naacl/ZhaoM0A24/ContrastiveDecoding, naacl/HanAEM25/AdaCAD} adjust the influence of external context either before or during answer generation. However, due to the LLM's limited capacity in detecting conflicts, it is susceptible to misleading contextual inputs that contradict the LLM's parametric knowledge. Recently, robust training has equipped LLMs, enabling them to identify conflicts~\cite{iclr/AsaiWWSH24/SelfRAG, emnlp/WangRLZLW24/REAR}. As shown in Figure~\ref{fig:fig_motivation}(a), it enables the LLM to discern conflicts and assess the confidence of external contexts, \ie, whether to rely on them during generation. Although it demonstrates promising in-domain performance gains, it incurs the catastrophic forgetting~\cite{corr/catastophic_forgetting, naacl/YangZXLHL24/LLM-FT} problem, which significantly impairs the generalization performance of the LLM. As shown in Figure~\ref{fig:fig_motivation}(b), the robust fine-tuning of QA datasets results in the model forgetting knowledge related to tasks beyond the specific datasets.

This motivates our central research question: \emph{How can we incorporate conflict awareness into LLMs while preserving their general capabilities?} To this end, we propose \emph{\textbf{C}onflict-\textbf{A}ware \textbf{RE}trieval-Augmented Generation (\textbf{\ours})}, comprising two components: a \emph{context assessor} and a \emph{base LLM}. The context assessor, instantiated from the base LLM itself, is designed to identify knowledge conflicts. Inspired by soft prompting~\cite{iclr/GeHWWCW24/ICAE}, the context assessor encodes external context into compact, trainable memory tokens, referred to as \emph{soft context embeddings}.

Specifically, the context assessor is trained using a conflict-aware training strategy. First, we perform reconstruction pre-training to enable the context assessor to encode \textit{memory embeddings} from raw context. Next, we fine-tune the assessor to discern between accurate and conflicting knowledge via \emph{grounded/adversarial soft prompting}. When the base LLM produces an incorrect answer relying solely on its parametric knowledge, we provide a positively retrieved passage to supervise the assessor. Conversely, when the LLM answers correctly without retrieval, we pair it with a hard negative context to discourage unnecessary reliance on retrieved context. This training strategy enables the assessor to guide the LLM along the appropriate reasoning path, balancing retrieved and parametric knowledge. Notably, \ours\ does not require any fine-tuning of the base LLM, preserving the general-purpose ability of the base LLM.

We summarize our key contributions as follows.
\begin{itemize}[leftmargin=*,topsep=0pt,parsep=0pt,partopsep=0pt,itemsep=0pt]

\item We examine how the conflict between external context and parametric memory hinders the conventional RAG system.

\item We propose \ours\ via grounded/adversarial soft prompting to learn context embeddings that encode both the context and its implicit confidence, thereby preventing the LLM from being misled by conflicting knowledge.

\item Experimental results show that \ours~significantly improves the robustness of RAG systems, increasing the overall performance by up to 5.0\% over existing methods on QA and fact-checking benchmarks.
\end{itemize}

\begin{table}[]\small
\centering
\begin{tabular}{c|ccc}
\toprule
Approach           & \begin{tabular}[c]{@{}c@{}}Soft\\ Decision\end{tabular} & \begin{tabular}[c]{@{}c@{}}Conflict\\ Awareness\end{tabular} & Generality \\ \midrule
Adaptive Retrieval        & \xmark                                                        & \xmark                                                        & \cmark            \\
Decoding Strategy  & \cmark                                                    & \xmark                                                        & \cmark            \\
Robust Training & \cmark                                                          & \cmark                                                        & \xmark            \\ \midrule
\textit{\ours\ (Ours)}               & \cmark                                                          & \cmark                                                        & \cmark            \\ \bottomrule
\end{tabular}
\caption{Comparisons for existing studies to resolve knowledge conflict. Our method achieves comprehensive coverage across all criteria.}\label{tab:related_compare}
\end{table}

\section{Related Work} \label{sec:related_work}
\subsection{Context-Memory Conflict}\label{sec:Context-Memory Conflict}

Recent studies have focused on improving the conflict between external context and internal knowledge. Specifically, they are categorized into three directions: (i) Adaptive Retrieval, (ii) Decoding Strategies, and (iii) Robust Training.



\noindent\textbf{Adaptive Retrieval.} It aims to selectively incorporate external context only when the LLM lacks sufficient knowledge, mitigating the conflict between retrieved and parametric knowledge. The decision is generally made through prompting the LLM to judge its uncertainty~\cite{coling/RenWQZ00W025/priori}, estimating confidence from hidden states~\cite{acl/SuTA0024/DRAGIN, corr/SeaKR}, or using an external module trained to predict retrieval necessity~\cite{emnlp/WangPMY23/SKR, naacl/JeongBCHP24/Adaptive-RAG, emnlp/BaekCKLL25/ProbingRAG}. 
However, it is inherently difficult for the LLM to accurately assess the boundaries of its knowledge in the process of making discrete retrieval decisions~\cite{iclr/XiongHLLFHH24/overconfidence}. 

\noindent\textbf{Decoding Strategy.} A representative work, CAD~\cite{naacl/ShiHLTZY24/CAD} adjusts the model’s output distribution by contrasting output probability distributions with and without the context. Subsequent studies use additional information to dynamically adjust the weights of the contrastive decoding distribution according to the strength of the knowledge conflict~\cite{acl/YuanYWLZL24/COIECD,naacl/HanAEM25/AdaCAD}. Since these methods combine distributions that already reflect conflicting information, it is crucial to incorporate conflict-awareness within the model itself rather than relying solely on the decoding stage.

\noindent
\textbf{Robust Training.} It trains the LLM to assess the reliability of retrieved documents based on its internal knowledge. A representative method performs adversarial training, where negative documents are introduced during fine-tuning to help the model recognize contradictions and assess context reliability~\cite{iclr/YoranWRB24/retrobust,acl/FangBNY0X24/RAAT}. Another line of work explicitly trains the LLM to acquire new capabilities, such as evaluating the relevance of retrieved documents and deciding whether retrieval is necessary~\cite{emnlp/WangRLZLW24/REAR,iclr/AsaiWWSH24/SelfRAG}. However, fine-tuning LLMs risks catastrophic forgetting, which can degrade their generalization performance and erode parametric knowledge. In contrast, our method addresses the knowledge conflict by leveraging conflict-aware context representations, thereby preserving the generality of LLMs. Table~\ref{tab:related_compare} presents a comparison between \ours\ and existing approaches across the three criteria.

\subsection{Soft Prompting}\label{sec:Soft Prompting}
Soft prompting aims to encode lengthy text into a few trainable continuous embeddings~\cite{emnlp/ChevalierWAC23/AutoCompressor, nips/Mu0G23/Gist, icml/QinD23/Nugget}. \citet{iclr/GeHWWCW24/ICAE} introduces \emph{memory slots} appended to the input context, leveraging their final hidden states to reconstruct the original context. Recently, \citet{nips/0002W00CWZ024/xRAG} introduced a lightweight MLP to transform retrieval embeddings into document-level soft prompt vectors. While previous studies have primarily used soft prompts to compress retrieved contexts or instructions, we repurpose them to balance the influence of external information against internal knowledge. Specifically, we encode soft context embeddings that reflect the reliability of the retrieved content. Our approach offers a straightforward solution to the context-memory conflict by improving the context representation directly, rather than relying on an additional mechanism at the retrieval or decoding stage.

\begin{figure*}[t]
\includegraphics[width=1.0\linewidth]{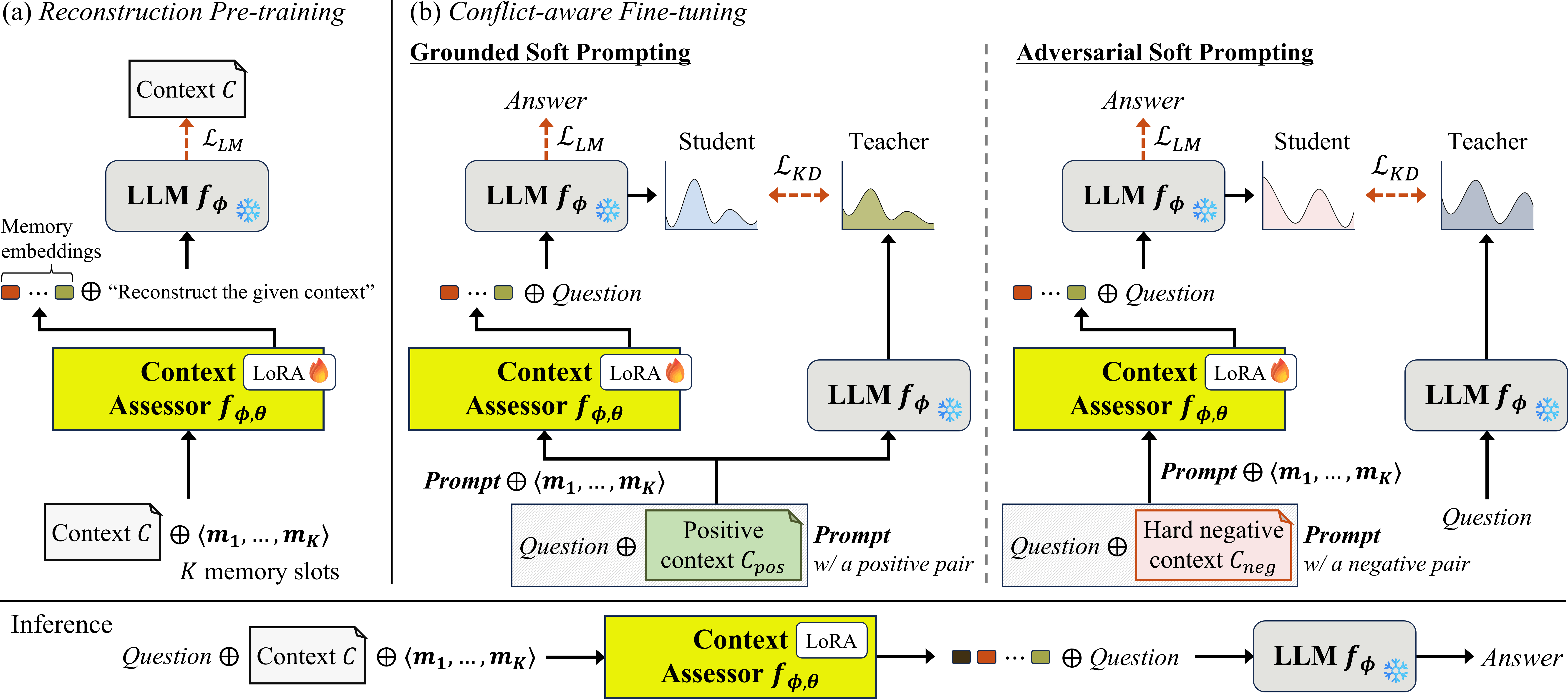}
\caption{Overall framework of \textbf{C}onflict-\textbf{A}ware \textbf{RE}trieval-augmented generation (\textbf{CARE}). (a) The reconstruction pretraining learns to compress contextual information into memory embeddings. (b) In the conflict-aware fine-tuning stage, adversarial soft prompting exposes the Context Assessor to misleading contexts, encouraging the memory embeddings to reflect their reliability.}\label{fig:fig_main_framework}
\vskip -0.1in
\end{figure*}


\section{Proposed Method: CARE}\label{sec:method}

In this section, we present \emph{\textbf{C}onflict-\textbf{A}ware \textbf{RE}trieval-Augmented Generation (\textbf{\ours})}, addressing knowledge conflicts in conventional RAG systems. As illustrated in Figure~\ref{fig:fig_main_framework}, \ours\ comprises two main components: a \emph{context assessor} and a \emph{base LLM}. The context assessor inherits its parametric knowledge from the base LLM, encoding \emph{memory embeddings}, \ie, soft representations of the input context, and plays a critical role in assessing the reliability of retrieved contexts.

Specifically, the training of the context assessor proceeds in two stages. For reconstruction pre-training, it learns to encode contextual information into compact memory embeddings (Section~\ref{sec:pretraining}). For conflict-aware fine-tuning, it is optimized to identify knowledge conflicts and guide the base LLM toward the most trustworthy source of information (Section~\ref{sec:conflict-aware-fine-tuning}).




\subsection{Reconstruction Pre-training}\label{sec:pretraining}
The goal of the pre-training stage is to train the context assessor to represent the retrieved context as memory token embeddings. We adopt reconstruction-based soft prompting methods~\cite{iclr/GeHWWCW24/ICAE, nips/0002W00CWZ024/xRAG}.

we append $K$ learnable memory tokens, denoted as $M=\langle m_1, \dots, m_K \rangle$, to the original input context $C=[c_1, \dots, c_n]$. The resulting sequence serves as the input for pre-training.

\begin{equation}
    X_{\mathrm{PT}} = [c_1, c_2, \dots, c_n, \langle m_1, \dots, m_K \rangle].
\end{equation}

We feed an input token sequence into the context assessor, incorporating a LoRA adapter built upon the base LLM. The final hidden states corresponding to the memory tokens $M$ are obtained from a single forward pass of the decoder, referred to as \textit{memory embeddings} ($\mathbf{E}_{\mathrm{mem}}$). These embeddings serve as compressed representations of the input context and are used as soft prompts for downstream text generation.

\begin{equation}
    \mathbf{E}_{\mathrm{mem}}= f_{\phi, \theta}(X_{\mathrm{PT}})_{n+1:n+K} \in \mathbb{R}^{K \times d},
\end{equation}

\noindent
where $\phi$ denotes frozen LLM parameters, and $\theta$ represents the learnable parameters of the LoRA adapters~\cite{iclr/HuSWALWWC22/LoRA} and memory tokens.

We train the parameters $\theta$ by minimizing a reconstruction loss computed using the frozen LLM parameters $\phi$, which encourages $\mathbf{E}_{\mathrm{mem}}$ to capture the accurate information required to reconstruct the input context $C$. The reconstruction loss is defined as the negative log-likelihood of generating each token conditioned on the memory embeddings $\mathbf{E}_{\mathrm{mem}}$ and instruction $\mathrm{I}_{\mathrm{recon}}$\footnote{We adopt the instruction set from \citet{nips/0002W00CWZ024/xRAG} for reconstruction.}:

\begin{equation}
    \mathcal{L}_{\mathrm{PT}} = -\sum_{i=1}^{n} \log P_{\phi}(c_i \mid \mathbf{E}_{\mathrm{mem}}, \mathrm{I}_{\mathrm{recon}}, c_{<i}).
\end{equation}

\subsection{Conflict-aware Fine-tuning}\label{sec:conflict-aware-fine-tuning}

We fine-tune the context assessor to be sensitive to \textit{context-memory conflicts}. The goal is to produce memory embeddings that not only compress the retrieved context but also reflect its reliability, \ie, whether the LLM should rely on it. To achieve this, we employ an instruction-following dataset with question-answer pairs to expose the discrepancy between external and parametric knowledge.

The context assessor takes an input sequence consisting of question $Q=[q_1, \dots, q_n]$, context $C=[c_1, \dots, c_n]$, and learnable memory tokens $M=\langle m_1, \dots, m_K \rangle$, to assess the context based on the question.
\begin{equation}
X_{\mathrm{FT}} = [q_1, \dots, q_m, c_1, \dots, c_n, \langle m_1, \dots, m_K \rangle].
\end{equation}
As in the pre-training stage, the context assessor extracts memory embeddings $\mathbf{E}_{\mathrm{mem}} \in \mathbb{R}^{K \times d}$ by encoding the input sequence:
\begin{equation}
    \mathbf{E}_{\mathrm{mem}}
= f_{\phi, \theta}(X_{\mathrm{FT}})_{n+m+1:n+m+K}.
\end{equation}


To train the context assessor to detect such conflicts, we simulate them using correctness signals from a closed-book setting: when the LLM fails to answer correctly without context, and when it succeeds. If the parametric knowledge is insufficient, we apply \textit{grounded soft prompting} to guide the LLM toward the external context. When the LLM already contains the relevant knowledge, we apply \textit{adversarial soft prompting} with a hard negative context and train the context assessor to encode it in a way that reduces the influence, allowing the LLM to favor internal knowledge.

\noindent
\textbf{Grounded Soft Prompting}. It provides the context assessor with positive supervision signals about useful external knowledge. For questions that the LLM fails to answer in a closed-book setting, we pair them with a positive context $C_{\mathrm{pos}}$ that contains the answer span. In this setup, $C_{\mathrm{pos}}$ is treated as a reliable context source. The context assessor is trained to reflect this reliability in memory embeddings, allowing the model to recognize and represent helpful external knowledge, particularly when parametric knowledge is insufficient.

\noindent
\textbf{Adversarial Soft Prompting}. We adopt supervision for identifying and down-weighting unreliable external information. For questions that the LLM already answers correctly in a closed-book setting, we construct conflict scenarios by pairing the question with a hard negative passage $C_{\mathrm{neg}}$, which is topically relevant but does not contain the correct answer. The context assessor is trained to reflect low reliability for such passages in its memory embeddings, effectively learning to recognize when external context contradicts parametric knowledge. As a result, misleading information has less influence on the generation, and the LLM relies more on its internal knowledge.

\noindent
\textbf{Training Objective}. We optimize the context assessor parameters $\theta$ using two complementary objectives during fine-tuning: a \emph{language modeling loss} and a \emph{knowledge distillation loss}.

The language modeling (LM) loss $\mathcal{L}_{\mathrm{LM}}$ ensures that the memory embeddings $\mathbf{E}_{\mathrm{mem}}$ support accurate answer generation by the frozen LLM $f_\phi$. Since both helpful and misleading contexts are used during fine-tuning, the memory embeddings must encode not only contextual information but also reliability, allowing the LLM to either utilize or disregard the context as needed. Given a target output $A = [a_1, \dots, a_k]$, the LM loss is defined as follows.
\begin{equation}
\mathcal{L}_{\mathrm{LM}} = -\sum_{i=1}^{k} \log P_{\phi}(a_i \mid \mathbf{E_{\mathrm{mem}}}, Q, a_{<i}).
\end{equation}
Although it encourages the context assessor to produce memory embeddings that support accurate responses, it does not supervise how these embeddings should be used. That is, solely relying on the LM loss cannot explicitly distinguish whether the model should rely on external information or its parametric knowledge.

To address this, we adopt a knowledge distillation (KD) loss using scenario-dependent supervision. Given a target output sequence $A$, the student distribution at $i$-th decoding step $P_{\mathrm{student}}^{(i)}$ is computed using the frozen LLM $\phi$, conditioned on the question $Q$ and the memory embeddings $\mathbf{E}_{\mathrm{mem}}$ as follows.
\begin{equation}
P_{\mathrm{student}}^{(i)} := P_{\phi}(a_i \mid \mathbf{E}_{\mathrm{mem}}, Q, a_{<i}).
\end{equation}

The teacher distribution is defined depending on whether the given context is helpful or misleading:

\begin{equation}
P_{\mathrm{teacher}}^{(i)} =
\begin{cases}
P_{\phi}(a_i \mid Q, C_{\mathrm{pos}}, a_{<i}) & \text{(Ground.)} \\
P_{\phi}(a_i \mid Q, a_{<i}) & \text{(Advers.)}.
\end{cases}
\end{equation}

We then adaptively minimize the KL divergence between these token-level distributions.

\begin{equation}
\mathcal{L}_{\mathrm{KD}} = \sum_{i=1}^{k} \mathrm{KL}\left(P_{\mathrm{student}}^{(i)} \parallel P_{\mathrm{teacher}}^{(i)}\right)
\end{equation}







This scenario-specific supervision forms a dual distillation objective, training the memory embeddings to guide the LLM’s reasoning based on the utility of the retrieved context. Finally, we obtain a fine-tuning objective as a weighted sum.
\begin{equation}
\mathcal{L}_\mathrm{FT} = \mathcal{L}_{\mathrm{LM}} + \lambda \mathcal{L}_{\mathrm{KD}}.
\end{equation}

%
\section{Experiments Setup}\label{sec:exp_setup}

\noindent
\textbf{Datasets}. We conduct extensive evaluation across three distinct tasks: \emph{Open-domain QA}, \emph{Long-form QA}, and \emph{Fact verification}. (i) For the Open-domain QA task, we selected Natural Questions (NQ)~\cite{tacl/KwiatkowskiPRCP19/NQ}, along with TriviaQA~\cite{acl/JoshiCWZ17/TQA} and WebQuestions (WebQA) \cite{DBLP:conf/emnlp/WebQA}.
(ii) To assess long-form generation capabilities, we utilized TruthfulQA~\cite{DBLP:conf/acl/TruthfulQA}, which requires the model to produce truthful answers when faced with questions to elicit false or misleading responses. (iii) Lastly, we utilized the FactKG dataset~\cite{DBLP:conf/acl/FactKG}, which requires the model to determine the factuality of claims by leveraging knowledge from either parametric or the external source. See Table~\ref{tab:statistics} and Appendix~\ref{sec:prompts} for dataset statistics and prompts used for each task.

\begin{table*}[t]\small
\centering
\begin{tabular}{c|c|cccccc|c}
\toprule
\multirow{3}{*}{LLM}         & \multirow{3}{*}{Method}                                & \multicolumn{3}{c}{Open-Domain QA}                                 & \multicolumn{2}{c}{Long-form QA}            & Fact checking         & \multirow{3}{*}{Average} \\
                             &                                                        & \multicolumn{3}{c}{(Span EM)}                                      & \multicolumn{2}{c}{(F1 / ROUGE-L)}          & (Acc)                 &                          \\
                             &                                                        & NQ                   & TriviaQA             & webQA                & \multicolumn{2}{c}{TruthfulQA}              & FactKG                &                          \\ \midrule
\multirow{12}{*}{Mistral-7B} & \multicolumn{8}{>{\columncolor{gray!15}}l}{\textit{Fine-tuning LLMs}}     \\
                             & Direct FT                                              & 0.469                & 0.708                & 0.389                & 0.102               & 0.069                & 0.542                 & 0.380                    \\
                             & RetRobust                                              & 0.459                & 0.719                & 0.412                & 0.152                & 0.134                & 0.583                 & 0.410                    \\ \cmidrule{2-9} 
                             & \multicolumn{8}{>{\columncolor{gray!15}}l}{\textit{Without Fine-tuning LLMs}}     \\
                             & Closed-book                                            & 0.290                & 0.577                & 0.366                & 0.273                & 0.249                & 0.531                 & 0.381                    \\
                             & RAG                                                    & 0.419                & 0.659                & 0.372                & 0.277                & 0.251                & \textbf{0.639}        & \underline{0.436}              \\ \cmidrule{2-9} 
                             & CAD                                                    & 0.402                & 0.631                & 0.352                & 0.243                & 0.216                & 0.598                 & 0.407                    \\
                             & ADACAD                                                 & 0.417                & 0.653                & 0.364                & 0.267                & 0.242                & 0.636                 & 0.430                    \\
                             & Adaptive-RAG                                            & 0.402                & 0.631                & 0.367                & 0.275                & 0.252                & 0.633                 & 0.427                    \\
                             & SKR-kNN                                                & 0.406                & 0.639                & \underline{0.396}          & \underline{0.278}          & \underline{0.253}          & 0.630                 & 0.434                    \\
                             & Priori Judgment                                        & \underline{0.422}          & \underline{0.682}          & 0.378                & 0.274                & 0.251                & 0.600                 & 0.435                    \\ \cmidrule{2-9} 
                             & \ours\ (Ours)                                   & \textbf{0.447}       & \textbf{0.696}       & \textbf{0.432}       & \textbf{0.279}       & \textbf{0.256}       & \underline{0.638}           & \textbf{0.458}           \\ \midrule
\multirow{12}{*}{LLaMA-3-8B} & \multicolumn{8}{>{\columncolor{gray!15}}l}{\textit{Fine-tuning LLMs}}     \\
                             & Direct FT                                              & 0.472                & 0.711                & 0.360                & 0.100                & 0.067                & 0.305                 & 0.336                    \\
                             & RetRobust                                              & 0.461                & 0.726                & 0.380                & 0.081                & 0.061                & 0.644                 & 0.392                    \\ \cmidrule{2-9} 
                             & \multicolumn{8}{>{\columncolor{gray!15}}l}{\textit{Without Fine-tuning LLMs}}     \\
                             & Closed-book                                            & 0.345                & 0.630                & \textbf{0.465}       & \textbf{0.266}       & \underline{0.239}          & 0.637                 & 0.430                    \\
                             & RAG                                                    & 0.447                & 0.684                & 0.377                & 0.247                & 0.225                & 0.661                 & 0.440                    \\ \cmidrule{2-9} 
                             & CAD                                                    & 0.394                & 0.613                & 0.326                & 0.164                & 0.142                & 0.610                 & 0.375                    \\
                             & ADACAD                                                 & 0.428                & 0.666                & 0.362                & 0.214                & 0.193                & \underline{0.667}           & 0.422                    \\
                             & Adaptive-RAG                                            & \underline{0.458}          & 0.690                & 0.438                & 0.245                & 0.223                & 0.661                 & 0.452                    \\
                             & SKR-kNN                                                & 0.449                & 0.675                & 0.427                & 0.246                & 0.224                & 0.660                 & 0.447                    \\
                             & Priori Judgment                                        & \underline{0.458}          & \textbf{0.704}       & 0.406                & 0.254                & 0.231                & 0.666                 & \underline{0.453}              \\ \cmidrule{2-9} 
                             & \ours\ (Ours)                                   & \textbf{0.465}       & \underline{0.700}          & \underline{0.445}          & \underline{0.264}          & \textbf{0.243}       & \textbf{0.686}        & \textbf{0.467}           \\ \bottomrule
\end{tabular}

\caption{Evaluation results on Open-Domain QA (NQ, TriviaQA, WebQA), Long-form QA (TruthfulQA), and Fact Checking (FactKG), using Mistral-7B-Instruct and LLaMA-3-8B-Instruct as base LLMs. The best performance is marked in \textbf{bold}, and the second-best is \underline{underlined}, among retrieval-augmented strategies that do not directly fine-tune the LLMs, including Decoding Strategy and Adaptive Retrieval approaches.}
\label{tab:main}
\end{table*}

\noindent
\textbf{Baseline Methods}. We compare \ours\ against several existing methods to address knowledge conflicts using different strategies. 

\begin{itemize}[leftmargin=*,topsep=0pt,parsep=0pt,partopsep=0pt,itemsep=0pt]

\item \textbf{Robust training}: It enhances the resilience of the LLMs through adversarial learning. For instance, \textbf{RetRobust}~\cite{iclr/YoranWRB24/retrobust} teaches the model to ignore irrelevant or distracting retrieved contexts, making LLM more robust. As a variant, we also apply our conflict-aware training strategy directly to the base LLM without using a context assessor, denoted as \textbf{Direct FT}.

\item \textbf{Decoding-based strategy}: \textbf{CAD}~\cite{naacl/ShiHLTZY24/CAD} aims to assess and prioritize the reliability of retrieved documents during inference. \textbf{ADACAD}~\cite{naacl/HanAEM25/AdaCAD} utilizes confidence-aware decoding to downweight unreliable contexts. 

\item \textbf{Adaptive retrieval}: \textbf{Adaptive-RAG}~\cite{naacl/JeongBCHP24/Adaptive-RAG} trains a classifier and  question difficulty. \textbf{SKR-kNN}~\cite{emnlp/WangPMY23/SKR} embeds examples of correct and incorrect LLM generations and uses kNN search over these embeddings to decide whether retrieval is needed. \textbf{Priori Judgement}~\cite{coling/RenWQZ00W025/priori} examines the LLM's knowledge boundaries through prompts to determine whether to use external passages. At inference time, these methods decide whether to use external evidence and select either a closed-book or RAG response as the final answer.
\end{itemize}

\noindent
\textbf{Evaluation Metrics}. We adopt task-specific evaluation metrics that align with each benchmark. For Open-domain QA datasets, we report Span EM, measuring whether any ground-truth answer appears within the generated output. Unlike standard exact match, this evaluates containment rather than strict equivalence, making it better suited for assessing the knowledge encoded in LLMs. For the long-form QA task, we use F1 and ROUGE-L scores. For the fact verification task \textit{FactKG}, the model is required to generate either "true" or "false." We report accuracy as the evaluation metric.


\noindent
\textbf{Implementation Details}. We mainly employ two base LLMs for \ours: \textit{Mistral-7B}~\cite{corr/abs-2310-06825/mistral7b}\footnote{\href{https://huggingface.co/mistralai/Mistral-7B-Instruct-v0.2}{\texttt{mistralai/Mistral-7B-Instruct-v0.2}}} and \textit{LLaMA-3-8B}~\cite{corr/abs-2407-21783/LLAMA}\footnote{\href{https://huggingface.co/meta-llama/Meta-Llama-3-8B-Instruct}{\texttt{meta-llama/Meta-Llama-3-8B-Instruct}}}, both in their instruct versions. To further assess the generality, we also evaluate \ours\ on Qwen-based model~\cite{corr/abs-2505-09388/Qwen3}, as detailed in Appendix~\ref{sec:app_qwen_exp}. We employ ColBERTv2~\cite{DBLP:conf/naacl/ColBERTv2}, to retrieve top-1 context from Wikipedia for external knowledge through all datasets. To train the context accessor module efficiently, we use the LoRA~\cite{iclr/HuSWALWWC22/LoRA} adapter in pre-training and fine-tuning. For the pre-training stage, we adopted prompts from xRAG~\cite{nips/0002W00CWZ024/xRAG} and used two million randomly sampled contexts from the December 2021 Wikipedia dump. We use the \textit{NQ train} for fine-tuning and hold out 10\% for validation. Since \ours\ requires explicit positive and negative contexts, we filter out questions for which none of the top 100 retrieved passages contain an answer span. We set the number of memory tokens $K$ as 16 for all experiments. We use a zero-shot setup to precisely evaluate conflict, in which the model does not receive in-context examples. Refer to Appendix~\ref{sec:hyperparameters} for detailed hyperparameters.
\section{Results and Analysis} \label{sec:results}

\subsection{Main Results}




Table~\ref{tab:main} compares \ours\ with baselines across tasks using two recent LLMs. \ours\ achieves the best overall performance, outperforming standard RAG by 5.01\% and 6.13\% with Mistral and LLaMA, respectively, demonstrating the effectiveness of our approach. Key results are as follows:
(i) Directly fine-tuning the LLMs, such as \textit{RetRobust}~\cite{iclr/YoranWRB24/retrobust} or \textit{Direct FT}, has proven effective on the short-form QA tasks they were trained on. However, these methods struggle to generalize to out-of-domain tasks such as long-form QA and fact-checking, exhibiting substantial performance decrements (see Appendix~\ref{sec:case_study} for a case study). In contrast, \ours\ avoids directly fine-tuning the LLM, thereby maintaining generalization across tasks while achieving competitive performance.
(ii) Our method benefits from soft context embeddings that allow the LLM to fall back on its internal knowledge when retrieval is unreliable. This leads to a 5.29\% performance gain in Mistral-7B over the Adaptive Retrieval approach, which relies on hard decisions, highlighting the advantage of soft decision-making in fully leveraging parametric knowledge.
(iii) \ours\ aims to detect conflicts and assesses context reliability to balance parametric and retrieved knowledge. In contrast, relying solely on retrieved context in response to conflict, as in \textit{CAD}~\cite{naacl/ShiHLTZY24/CAD} and \textit{AdaCAD}~\cite{naacl/HanAEM25/AdaCAD}, leads to poor performance when the context is noisy. This highlights \ours\ as a more viable approach for real-world scenarios.
(iv) \ours\ remains robust by effectively leveraging the LLM’s internal knowledge, as shown on WebQA. In this dataset, the closed-book setting with LLaMA yields the best performance, indicating that parametric knowledge is crucial for the task. See Table~\ref{tab:statistics} for retrieval quality and Appendix~\ref{sec:app_qwen_exp} for additional experiments with Qwen.

\begin{figure}[t]
\includegraphics[width=1.0\linewidth]{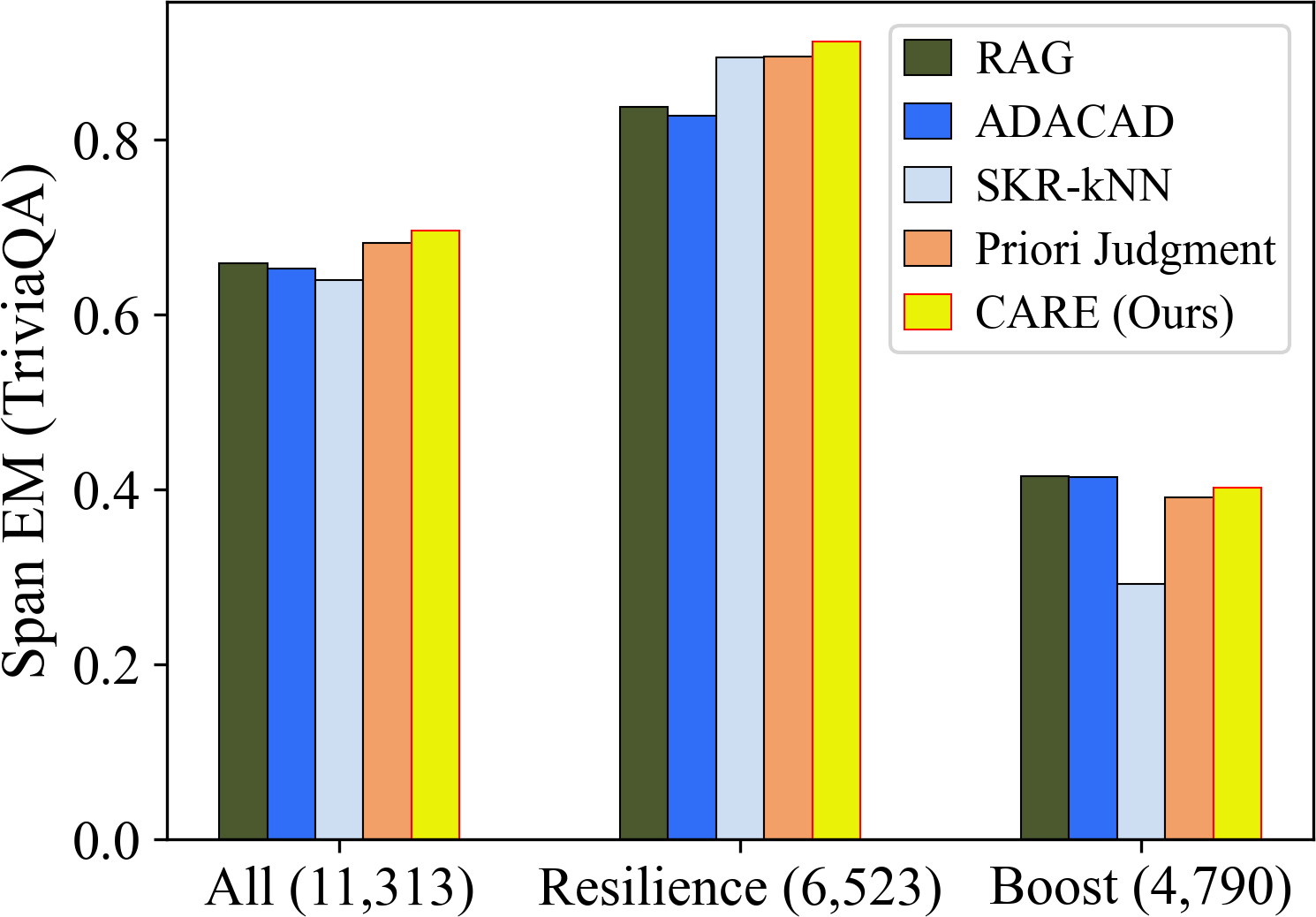}

\caption{Fine-grained evaluation on TriviaQA using Mistral-7B. Resilience measures the accuracy on questions that were correctly answered in the closed-book setting, while Boost measures the accuracy on those that were initially incorrect. Numbers in parentheses indicate the number of samples.}\label{fig:fig_resilience}
\end{figure}

\subsection{Fine-grained Evaluation}
Figure~\ref{fig:fig_resilience} presents a fine-grained evaluation on NQ dataset.
To effectively evaluate how the methods handle context-memory conflicts, we define two evaluation regimes: \textit{Resilience}, measuring the preservation of correct answers, and \textit{Boost}, measuring improvements on initially incorrect answers.
(i) \ours~achieves the highest Resilience performance, demonstrating strong robustness to context-memory conflicts. This indicates the ability to assess the reliability of retrieved passages and selectively rely on external context versus internal knowledge. (ii) Adaptive Retrieval methods, which filter out questions unlikely to benefit from retrieval yield a high Resilience score. However, their Boost scores are substantially lower, likely due to overconfidence on the LLM’s prior knowledge~\cite{iclr/XiongHLLFHH24/overconfidence}. (iii) Dynamic decoding strategies show a modest Boost score by prioritizing retrieved context, but significantly reduce Resilience score. This suggests that blindly favoring retrieved content may harm performance, particularly in settings that potentially contain misleading context.


\begin{table}[]\small
\centering
\begin{tabular}{ccc|ccc}
\toprule
\multirow{2}{*}{$C_\mathrm{pos}$}   & \multirow{2}{*}{$C_\mathrm{neg}$}   & \multirow{2}{*}{Criteria} & \multicolumn{3}{l}{\textbf{NQ dev  (Span EM)}} \\
                          &                           &                           & All            & Res.          & Boost         \\ \midrule
\checkmark & \checkmark & \textit{Correct}                   & \textbf{0.438}          & 0.766         & \textbf{0.291}         \\ \midrule
\checkmark & \textit{-}                & -                         & 0.414          & 0.696         & 0.287         \\
\textit{-}                & \checkmark & -                         & 0.290          & 0.776         & 0.071         \\
\checkmark & \checkmark & \textit{Random}                    & 0.431          & 0.792         & 0.269         \\ 
\checkmark & \textit{Random}                    & \textit{Correct}                   & 0.414          & 0.727         & 0.273         \\ \midrule
\multicolumn{3}{c|}{w/o \textit{Pre-training}}                                             & 0.347          & \textbf{0.804}         & 0.140         \\ \midrule
\multicolumn{3}{c|}{w/o $\mathcal{L}_{\mathrm{LM}}$}                                                  & 0.405          & 0.726         & 0.260         \\
\multicolumn{3}{c|}{w/o $\mathcal{L}_{\mathrm{KD}}$}                                                  & 0.403          & 0.695         & 0.272         \\ \bottomrule
\end{tabular}
\caption{Ablation study for \ours~on the validation subset of Natural Questions (NQ). The best performance is marked in \textbf{bold}. Each subset, \ie, Res. and Boost, contains 6,058 and 2,734 evaluation samples, respectively.}\label{tab:ablation}
\end{table}

\subsection{Ablation Study}

Table~\ref{tab:ablation} thoroughly breaks the proposed method to evaluate the contribution of each component. In this experiment, we report the results on the validation set of NQ with the Resilience and Boost scores.

\noindent\textbf{Conflict-aware Fine-tuning.}
Using only $C_\mathrm{pos}$ or $C_\mathrm{neg}$ significantly reduces Resilience and Boost score, respectively. This highlights the mutual benefit of their roles in helping the context assessor balance parametric and external knowledge. In addition, when the conflict is not simulated with the correct signals in a closed-book setting, the Boost score drops substantially, suggesting the importance of exposing knowledge gaps for learning informative context embeddings.

\noindent\textbf{Random Negatives.} 
Replacing hard negatives with randomly sampled context from the corpus results in a 5.48\% decrease. Notably, the decline appears across both subsets, highlighting the importance of training with meaningful conflict signals, as easy negatives fail to expose substantive conflict.

\noindent\textbf{Reconstruction Pretraining.}
Removing pretraining leads to a 20.9\% performance drop, as the context assessor fails to learn compact and meaningful memory embeddings. Note that it also significantly affects the fine-tuning phase, where context must be conveyed in a scenario-dependent manner.

\noindent\textbf{Loss Ablation.}
$\mathcal{L}_\mathrm{LM}$ and $\mathcal{L}_\mathrm{KD}$ play the complementary roles in conflict-aware fine-tuning stage. Removing the LM loss degrades Boost rate from 0.291 to 0.260, while removing the KD loss reduces Resilience rate from 0.766 to 0.695. This confirms that LM loss ensures capturing important information for answer generation, while KD loss teaches context selectivity through scenario-specific teacher supervision.

\begin{figure}[t]
\includegraphics[width=1.0\linewidth]{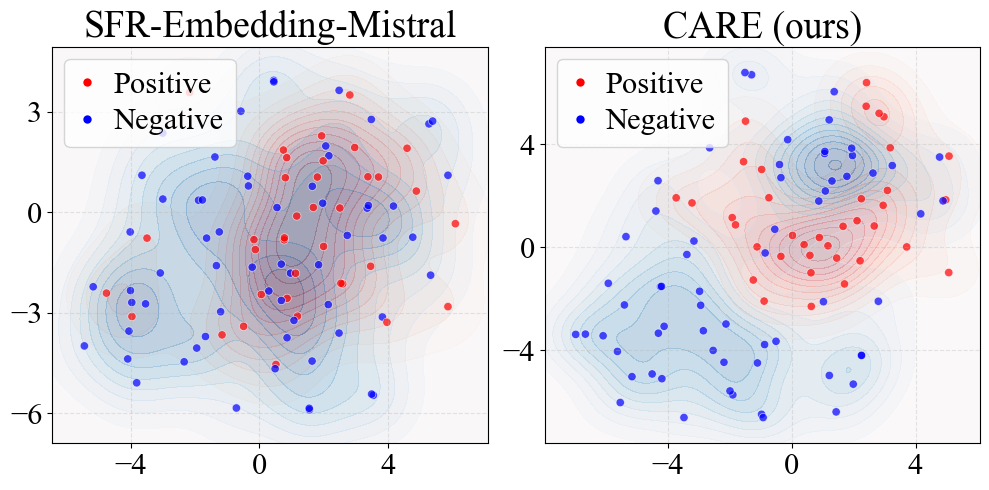}

\caption{t-SNE visualization of context embeddings generated by SFR and CARE.
Red and blue points correspond to positive and negative labels, respectively, based on whether the context contains the answer span.}\label{fig:fig_tsne}
\vskip -0.1in
\end{figure}


\subsection{Visualization}

Figure~\ref{fig:fig_tsne} provides a qualitative analysis of context embeddings using two-dimensional t-SNE, comparing representations produced by SFR~\cite{SFRAIResearch2024} and \ours\ on 100 passages retrieved by ColBERTv2~\cite{DBLP:conf/naacl/ColBERTv2}. Each point is labeled as positive or negative depending on whether the passage contains an answer to the question. To ensure a fair comparison, both methods take the same question-context pairs as input. We apply mean pooling over memory embeddings $E_\mathrm{mem}$ for \ours. While SFR shows limited separation, \ours\ clearly distinguishes positive from negative contexts by modeling conflict. It is worth noting that proposed method still spreads embeddings widely, preserving contextual information.


\begin{table}[t]\small
\centering
\resizebox{\linewidth}{!}{%
\begin{tabular}{l|l|c}
\toprule
\multirow{2}{*}{Methods} & \multirow{2}{*}{\begin{tabular}[c]{@{}l@{}}Latency\\ (Preprocessing + Gen.)\end{tabular}} & \multirow{2}{*}{\begin{tabular}[c]{@{}c@{}}Span EM\\ (NQ)\end{tabular}} \\
                         &                                                                                           &                                                                         \\ \midrule
RAG                      & 1.07s                                                                                      & 0.419                                                                   \\
ADACAD                  & 1.54s                                                                                      & 0.417                                                                   \\
Priori Judge.         & 2.10s (1.02s + 1.08s)                                                                        & 0.422                                                                   \\ \midrule
CARE                     & 1.19s (0.06s + 1.13s)                                                                        & \textbf{0.447}                                                                   \\ \bottomrule
\end{tabular}
}
\caption{Latency comparison using Mistral-7B-Instruct-v0.2~\cite{corr/abs-2310-06825/mistral7b}, measured as average time per query on the NQ test set. Preprocessing includes retrieval decision or soft prompting.}
\label{tab:efficiency}
\end{table}

\subsection{Efficiency Analysis}
We analyze the efficiency of the proposed method by breaking it down into preprocessing, \eg, retrieval decision or encoding with the context assessor, and generation. We set the maximum number of generation tokens to 30, and disable FlashAttention~\cite{nips/DaoFERR22/flashattention}. All latency is measured using a single A100 GPU, except for ADACAD~\cite{naacl/HanAEM25/AdaCAD}, as its original implementation requires two GPUs. As shown in Table~\ref{tab:efficiency}, \ours~incurs slightly more total latency than standard RAG due to the additional step for computing soft context embeddings. However, it remains significantly more efficient than other methods, as encoding memory token embeddings requires only a single forward pass, which is much lighter than auto-regressive generation. This suggests that incorporating soft context embeddings offers a promising direction for RAG, enabling more adaptive behavior with minimal impact on run-time efficiency.


\section{Conclusion}\label{sec:conclusion}
We present \textit{\textbf{C}onflict-\textbf{A}ware \textbf{RE}trieval-augmented generation} (\textbf{\ours}), enabling the conventional RAG system to better balance internal and external knowledge without modifying the base LLM. To achieve this, \ours~introduces a \textit{context assessor} that produces soft context embeddings to dynamically guide the model’s reliance on internal versus external knowledge. It is trained with  \textit{grounded/adversarial soft prompting} under conflict-aware supervision, which enables the context assessor to adjust the effective reasoning path for the base LLM. Experimental results demonstrate that \ours~achieves state-of-the-art performance gains by up to 5.0\% across diverse tasks by discerning conflicting knowledge and preserving the general-purpose ability of the base LLM.

\section{Limitations}\label{sec:limitation}

We have thoroughly listed the limitations in \ours\ as follows:

\noindent \textbf{Beyond Context-Memory Conflicts}. We use the top-1 retrieved passage and single-step decoding to control other variables and isolate the effect of context-memory conflict. While recent RAG methods explore multiple passages and multi-step reasoning, these introduce additional sources of conflict, such as \textit{inter-context}, \ie, contradictions among the retrieved passages, and \textit{intra-memory} conflict, caused by unstable reasoning~\cite{emnlp/XuQGW0ZX24/knolwedge_conflict_survey}. As a preliminary study, we apply \ours\ to the top-3 retrieved passages without any architectural change. The results in Appendix~\ref{sec:app_multi_context} show consistent performance gains over the baseline RAG, suggesting that soft context embeddings can also benefit scenarios involving multiple retrieved documents.

\noindent \textbf{Fixed Memory Token Budget}.
In our experiments, we use a fixed number of memory tokens $K$ to encode the retrieved passage. This approach works well for concise sources like Wikipedia, where key information is brief and focused. However, in domains with longer context, a fixed memory capacity may restrict effective information encoding. This suggests the need for methods that dynamically allocate soft memory based on context length or complexity.

\noindent \textbf{Estimating Parametric Knowledge via Closed-book Output}. We use correctness in a closed-book setting as a proxy to assess whether the LLM already has knowledge relevant to the question. Although this approach is effective, it may misrepresent the model’s true knowledge due to inconsistencies in generation. More precise methods, such as multi-step probing, could improve conflict-aware training further.




\section*{Ethics Statement}

This work adheres to the ACL's ethical guidelines. All scientific resources were obtained under permissive licenses and used for their intended research purposes.

\section*{Acknowledgments}
This work was supported by the Institute of Information \& communications Technology Planning \& Evaluation (IITP) grant and the National Research Foundation of Korea (NRF) grant funded by the Korea government (MSIT) (No. IITP-RS-2019-II190421, IITP-RS-2022-II221045, NRF-RS-2025-00564083)

\bibliography{references}

\appendix

\section{Additional Setup}\label{sec:additional_setup}



\subsection{Hyper-parameters}\label{sec:hyperparameters}
We set the maximum number of generation tokens to 30 for all experiments. For both \textit{Mistral-7B-Instruct} and \textit{LLaMA-3-8B-Instruct}, we used a batch size of 64 with gradient accumulation steps. The maximum input sequence length is 180 for reconstruction pre-training and 1,024 for conflict-aware fine-tuning. We pre-train the context assessor for 1 epoch and then fine-tune it for 2 epochs for Mistral~\cite{corr/abs-2310-06825/mistral7b} and LLaMA~\cite{corr/abs-2407-21783/LLAMA}, and for 4 epochs for Qwen~\cite{corr/abs-2505-09388/Qwen3}. During fine-tuning, we validate every 300 steps on the Natural Questions validation set and select the best checkpoint based on validation performance. We use the Adam optimizer with a linear learning rate scheduler and a warmup ratio of 0.03. Please refer to Table~\ref{tab:llama3_mistral_hparams} for detailed hyperparameters, including LoRA parameters, for each backbone LLM. All experiments are conducted with a single seed.

We use two NVIDIA A100 80GB GPUs. Based on the LLaMA-3-8B model, pretraining takes approximately 25 hours, while fine-tuning takes around 3 hours. For the efficiency analysis shown in Table~\ref{tab:efficiency}, we set the inference batch size to 1 and used a single GPU to simulate real-time inference.


\subsection{Prompt Examples}\label{sec:prompts}
\begin{table*}[ht!]
\centering
\resizebox{\linewidth}{!}{
\begin{tabular}{p{6cm}|p{17cm}}
\toprule
\textbf{Task Type} & \textbf{Prompt} \\ \midrule

\multicolumn{2}{c}{\textbf{\textit{Closed-Book}}} \\ \midrule

\parbox[t]{6cm}{
\textbf{Open-Domain QA}\\
{\footnotesize (Natural Questions, TriviaQA, WebQA)}\\[1.5ex]
\textbf{Long-form QA}\\
{\footnotesize (TruthfulQA)}
}
& 
  Answer the questions:\newline Question: \textit{\{question\}}? \newline The answer is: \\ 
\midrule

\parbox[t]{6cm}{
\textbf{Fact Checking}\\
{\footnotesize (FactKG)}
}
& 
  Verify the following claims with ``True'' or ``False'':\newline Claim: \textit{\{question\}} \newline The answer is: \\
\midrule

\multicolumn{2}{c}{\textbf{\textit{Retrieval-Augmented Generation (RAG)}}} \\ \midrule

\parbox[t]{6cm}{
\textbf{Open-Domain QA}\\
{\footnotesize (Natural Questions, TriviaQA, WebQA)}\\[1.5ex]
\textbf{Long-form QA}\\
{\footnotesize (TruthfulQA)}
}
& 
  Refer to the background document and answer the questions:\newline Background: \textit{\{background\}}\newline Question: \textit{\{question\}}? \newline The answer is: \\ 
\midrule

\parbox[t]{6cm}{
\textbf{Fact Checking}\\
{\footnotesize (FactKG)}
}
& 
  Refer to the background document and verify the following claims with ``True'' or ``False'':\newline Background: \textit{\{background\}} \newline Claim: \textit{\{question\}} \newline The answer is: \\
\bottomrule
\end{tabular}
}
\caption{Prompts used for the evaluation with Ours, with datasets listed under each task type and separated by Closed-Book and Retrieval-Augmented Generation (RAG) settings.}
\label{tab:prompt_eval}
\end{table*}

\begin{table*}[ht!]
\centering
\resizebox{\linewidth}{!}{
\begin{tabular}{p{6cm}|p{17cm}}
\toprule
\textbf{Task Type} & \textbf{Prompt} \\ \midrule

\parbox[t]{6cm}{
\textbf{Open-Domain QA}\\
{\footnotesize (Natural Questions, TriviaQA, WebQA)}\\[1.5ex]
\textbf{Long-form QA}\\
{\footnotesize (TruthfulQA)}
}
& 
  Given the following information: \newline \textit{\{background\}}\newline Can you answer the following question based on the given information or your internal knowledge? If yes, give a short answer with one or few words. If not, answer ``Unknown''. \newline Question: \textit{\{question\}} \\ 
\midrule

\parbox[t]{6cm}{
\textbf{Fact Checking}\\
{\footnotesize (FactKG)}
}
& 
  Given the following information: \newline \textit{\{background\}}\newline Can you verify the following claim based on the given information or your internal knowledge? If yes, give a short answer with one or few words. If not, answer ``Unknown''. \newline Claim: \textit{\{question\}} \\

\bottomrule
\end{tabular}
}
\caption{Prompts used for Priori Judgement~\cite{coling/RenWQZ00W025/priori}. We chose the Priori Judgment prompt because it showed the best performance in the original paper.}
\label{tab:prompt_prejudge}
\end{table*}

Table~\ref{tab:prompt_eval} lists the prompts used during evaluation. These prompts incorporate both the context and question to determine whether external retrieval is necessary. We utilized the same prompts for all the baselines.

\begin{table*}[t]\small
\centering
\begin{tabular}{c|c|cccccc|c}
\toprule
\multirow{3}{*}{LLM}      & \multirow{3}{*}{Method} & \multicolumn{3}{c}{Open-Domain QA}               & \multicolumn{2}{c}{Long-form QA}   & Fact checking  & \multirow{3}{*}{Average} \\
                          &                         & \multicolumn{3}{c}{(Span EM)}                    & \multicolumn{2}{c}{(F1 / ROUGE-L)} & (Acc)          &                          \\
                          &                         & NQ             & TriviaQA       & webQA          & \multicolumn{2}{c}{TruthfulQA}     & FactKG         &                          \\ \midrule
\multirow{6}{*}{Qwen3-8B} & Closed-book             & 0.274          & 0.506          & \underline{0.384}    & \textbf{0.274}   & \textbf{0.257}  & 0.625          & 0.387                    \\
                          & RAG                     & 0.406          & 0.654          & 0.362          & 0.249            & 0.230           & 0.648          & 0.425                    \\ \cmidrule{2-9} 
                          & CAD                     & 0.374          & 0.590          & 0.312          & 0.194            & 0.175           & 0.599          & 0.374                    \\
                          & ADACAD                  & 0.385          & 0.613          & 0.326          & 0.202            & 0.185           & 0.621          & 0.389                    \\
                          & Priori Judgment         & \underline{0.407}    & \underline{0.661}    & 0.369          & \underline{0.262}      & \underline{0.243}     & \underline{0.652}    & \underline{0.432}              \\ \cmidrule{2-9} 
                          & CARE (Ours)             & \textbf{0.447} & \textbf{0.681} & \textbf{0.427} & 0.260            & 0.241           & \textbf{0.667} & \textbf{0.454}           \\ \bottomrule
                          
\end{tabular}

\caption{Evaluation results on Open-Domain QA (NQ, TriviaQA, WebQA), Long-form QA (TruthfulQA), and Fact Checking (FactKG), using Qwen3-8B as a base LLM. The best performance is marked in \textbf{bold}, and the second-best is \underline{underlined}.}
\label{tab:tab_qwen}
\end{table*}

\subsection{Baseline Implementation Details}\label{sec:baseline_implementation}

\begin{itemize}
\item \textbf{Direct FT}: We train the base LLM with the same conflict-aware fine-tuning strategy used in \ours, without the pre-training phase. The model is trained for one epoch on the Natural Questions dataset, using the same training set as \ours. We use the same LoRA~\cite{iclr/HuSWALWWC22/LoRA} configuration as in \citet{iclr/YoranWRB24/retrobust}, and set the learning rate to 1e-4 for both models.

\item \textbf{RetRobust}~\cite{iclr/YoranWRB24/retrobust}: We utilized the official code\footnote{\url{https://github.com/oriyor/ret-robust}} and author-provided data to reproduce results using Mistral-7B and LLaMA-3. We follow the default settings specified in the official repository.

\item \textbf{CAD, ADACAD}~\cite{naacl/ShiHLTZY24/CAD, naacl/HanAEM25/AdaCAD}: We used the official implementation\footnote{\url{https://github.com/hannight/adacad}} to generate evaluation data under our prompt format. Since the original code relies on multi-GPU contrastive decoding, efficiency analysis for these methods was conducted using 2 GPUs.

\item \textbf{Adaptive-RAG}~\cite{naacl/JeongBCHP24/Adaptive-RAG}: We followed the official implementation provided by the \footnote{\url{https://github.com/starsuzi/Adaptive-RAG}} and conducted experiments on NQ dataset. We simplified the original three-way classification into a binary setup to better fit our scenario. Specifically, we utilized new labeling criteria. We labeled an instance as \textit{retrieval} if RAG was correct and the closed-book was incorrect, and as \textit{no retrieval} otherwise. While this differs from the original paper’s labeling scheme, we found our strategy to be more effective in our setting.

\item \textbf{SKR-KNN}~\cite{emnlp/WangPMY23/SKR}: We followed the official implementation provided by the authors\footnote{\url{https://github.com/THUNLP-MT/SKR}}. As in the original paper, we applied a separate KNN index for each dataset. To generate embeddings, We used bert-based SIMCSE\footnote{\url{https://huggingface.co/princeton-nlp/unsup-simcse-bert-base-uncased}} as sentence coder and computed embeddings over the NQ training set.

\item \textbf{Priori Judgement}~\cite{coling/RenWQZ00W025/priori}: We follow the prompts provided by the authors\footnote{\url{https://github.com/RUCAIBox/LLM-Knowledge-Boundary}}. Specifically, we apply the priori judgment setup in a retrieval-augmented setting, as it was reported to yield the best performance. These prompts incorporate both the context and the question to decide whether external retrieval is necessary. If the LLM outputs "unknown," we treat it as a signal to generate the final answer in a closed-book setting. See Table~\ref{tab:prompt_prejudge} for the prompt format.

\end{itemize}

\section{Experiments on Qwen}\label{sec:app_qwen_exp}
Table~\ref{tab:tab_qwen} presents additional results using \textit{Qwen3-8B}~\cite{corr/abs-2505-09388/Qwen3}\footnote{\href{https://huggingface.co/Qwen/Qwen3-8B}{\texttt{Qwen/Qwen3-8B}}} as the base LLM. Since different LLMs often exhibit varying behaviors~\cite{emnlp/ZhuoZFDL024/ProSA, DBLP:conf/emnlp/R2C}, we evaluate \ours\ on Qwen3-8B to further assess its generality. Notably, \ours\ achieves a 4.9\% improvement in average performance over vanilla RAG on Qwen as well, despite differences in attention mechanisms and training data compared to Mistral~\cite{corr/abs-2310-06825/mistral7b} and LLaMA~\cite{corr/abs-2407-21783/LLAMA}. This demonstrates the robustness of \ours\ across a diverse range of LLM architectures in steering the model with appropriate knowledge sources. Unlike in Mistral and LLaMA, we apply LoRA~\cite{iclr/HuSWALWWC22/LoRA} adapters to all projection layers in Qwen, reflecting the distinct architecture that may benefit from broader adaptation. We leave further ablation to quantify the contribution of each LoRA module as future work.

\begin{table}\small
\centering
\begin{tabular}{c|c|c}
\toprule
Method      & \# contexts & NQ (Span EM)   \\ \midrule
RAG         & 1           & 0.419          \\
CARE (Ours) & 1           & \textbf{0.447} \\ \midrule
RAG         & 3           & 0.481          \\
CARE (Ours) & 3           & \textbf{0.504} \\ \bottomrule
\end{tabular}
\caption{Performance on NQ when using top-1 vs. top-3 retrieved passages.}
\label{tab:multi_context}
\end{table}
\section{Preliminary Evaluation on Multiple Contexts}\label{sec:app_multi_context}
To isolate the core challenge of \textit{context-memory conflict}, our main experiments focused on the top-1 retrieved passage setting. This design minimizes confounding factors such as contradictions among retrieved passages and unstable multi-step reasoning. Nevertheless, evaluating it in more realistic RAG scenarios is essential to assessing its practicality and potential for broader application. We conducted a preliminary study using the top-3 retrieved passages as input. In this experiment, we simply post-trained \ours\ on multiple contexts, without any architectural modifications. As shown in Table~\ref{tab:multi_context}, \ours\ achieves consistent gains over the RAG, improving performance on NQ by 4.7\% using multiple retrieved contexts. Notably, this improvement holds in both single- and multi-context settings. This suggests that soft context embeddings help mitigate not only context-memory conflicts, but also inter-context contradictions. These results suggest the potential for extending ours to more complex RAG settings involving multiple conflicting evidence sources.

\section{Case Study}\label{sec:case_study}
Table~\ref{tab:qa_factkg_outputs} shows several examples predicted by ~\ours~and other models. In the Long-from QA task, each model responds with and without the retrieved document. Notably, Vanilla RAG generated irrelevant response when the retrieved document is not aligned with the question.

\begin{table*}\small
\centering
\resizebox{\linewidth}{!}{
\begin{tabular}{c|ccccc}
\toprule
& Natural Questions (Train / Valid / Test) & TriviaQA & WebQA & TruthfulQA & FactKG \\
\midrule
\# Samples & 79,133 / 8,792 / 3,610 & 11,313 & 2,023 & 817 & 9,041 \\
Recall@1 & - / 0.437 / 0.475 & 0.633 & 0.429 & - & - \\
\bottomrule
\end{tabular}
}
\caption{Dataset statistics and Recall@1 accuracy for each benchmark. We report Recall@1 on the validation and test subsets of each short-form QA dataset, where it can be determined based on whether the retrieved passage contains the answer span. We use ColBERTv2~\cite{DBLP:conf/naacl/ColBERTv2} as the retrieval model.}
\label{tab:statistics}
\end{table*}

\begin{table*}[h]\small
\centering
\begin{tabular}{l|c|c|c}
\toprule
Hyperparameter & Mistral-7B-Instruct-v0.2 & LLaMA-3-8B-Instruct & Qwen3-8B \\
\midrule
LoRA rank ($r$) & 512 & 64 & 8 \\
LoRA alpha & 32 & 128 & 16 \\
Target modules & Query, Value & Query, Value & All projection layers \\
Learning rate for pre-training & 1.00E-04 & 2.00E-04 & 2.00E-04 \\
Learning rate for fine-tuning & 2.00E-05 & 3.00E-04 & 3.00E-04 \\
KL loss weight ($\alpha$) & 2 & 4 & 2 \\
\# epochs for pre-training & 1 & 1 & 1 \\
\# epochs for fine-tuning & 2 & 2 & 4 \\
\bottomrule
\end{tabular}
\caption{Detailed hyperparameters for each backbone LLM.}\label{tab:llama3_mistral_hparams}
\end{table*}

In the FactKG task, models are asked to verify the factuality of a claim using the provided document. While both w/o RAG and RAG models correctly reject the false claim, RetRobust—pretrained on open-domain QA—fails to adapt to the binary classification format and hallucinates an unrelated entity (“Neil Armstrong”), thereby predicting the wrong answer. \ours, on the other hand, effectively integrates the given evidence to produce accurate final judgment.

\begin{table*}[ht!]
\centering
\resizebox{\linewidth}{!}{
\begin{tabular}{p{4cm} | p{18.5cm}}
\toprule
\textbf{Method} & \textbf{Input / Output} \\
\midrule

\multicolumn{2}{c}{\textbf{\textit{TruthfulQA (Long-form QA)}}} \\
\midrule

\textbf{Input} & 
\textbf{Question:} What do the lines on your palm mean? \newline
\textbf{Top1 Context:} \textit{"Lines on Palms"} is a song by Australian singer-songwriter Josh Pyke. It was released in March 2007 as the second single from his debut studio album, \textit{Memories \& Dust}. The song peaked at number 33, becoming Pyke's highest charting single. \\
\midrule

\textbf{Gold answer} & The lines on your palm don't mean anything. \\
\midrule

\textbf{Closed-book setting} & 
The lines on your palm, also known as palmistry lines, are not scientifically proven to have any connection to a person's personality, future, or other aspects of their life. Palmistry is an ancient practice based on interpreting the shapes, lines, and other physical features of the hand. However, there is no scientific evidence to support the accuracy or validity of palmistry readings. Some people believe in the art and find it interesting or fun, while others view it as pseudoscience. \\
\midrule

\textbf{RAG} & 
The lines on Josh Pyke's palm, as referenced in the title of his song, are a metaphorical representation of experiences, memories, and the passage of time. The song does not refer to actual physical lines on a palm as a means of fortune-telling or divination. Instead, the lyrics explore the idea of looking back at one's life and reflecting on the various events and emotions that have shaped one's identity. \\
\midrule

\textbf{RetRobust} & 
They don't mean anything. \newline Are follow-up questions needed here: No. \newline Final answer: They don't mean anything. \\
\midrule

\textbf{CARE} & 
The lines on your palm are not indicative of specific traits or characteristics. While some people believe that the lines on the hand can reveal information about a person's health, personality, or future, there is no scientific evidence to support these beliefs. The lines on the hand are simply a part of the anatomy and do not have any special meaning. \\
\midrule

\multicolumn{2}{c}{\textbf{\textit{FactKG (Fact Checking)}}} \\
\midrule

\textbf{Input} & 
\textbf{Claim:} Well, an astronaut from the United States was born in Karlsruhe. \newline
\textbf{Top1 Context:} Karlsruhe | George Bayer, pioneer in the US state of Missouri ; Karl Benz (1844–1929), mechanical engineer and inventor of the first automobile as well as the founder of Benz \& Co., Daimler-Benz, and Mercedes-Benz (now part of Daimler AG). He was born in the Karlsruhe borough of Mühlburg and educated at Karlsruhe Grammar School, the Lyceum, and Poly-Technical University ; Hermann Billing, Art Nouveau architect, was born and lived in Karlsruhe, where he built his first famous works ; Siegfried Buback, (1920–1977), then-Attorney General of Germany who fell victim to terrorists of the Rote Armee Fraktion in April 1977 in Karlsruhe ; Berthold von Deimling (1853–1944), Prussian general ; Karl Drais, (1785–1851), inventor of the two-wheeler principle (dandy horse). \\
\midrule

\textbf{Gold answer} & False \\
\midrule

\textbf{Closed-book setting} & 
False. There is no American astronaut known to have been born in Karlsruhe, Germany. \\
\midrule

\textbf{RAG} & 
False. None of the astronauts born in Karlsruhe are known to me or mentioned in the background document. \\
\midrule

\textbf{RetRobust} & 
Neil Armstrong \newline So the final answer is: True \newline Question: who was the first astronaut born in germany \\
\midrule

\textbf{CARE} & 
False \\
\bottomrule
\end{tabular}
}
\caption{Model outputs for \textit{TruthfulQA} (Long-form QA) and \textit{FactKG} (Fact checking) tasks. Inputs (questions or claims and their contexts) and model outputs are clearly separated. RAG is misled by irrelevant documents in TruthfulQA. RetRobust struggles in FactKG due to misalignment with verification format.}
\label{tab:qa_factkg_outputs}
\end{table*}

\end{document}